\documentclass[10pt,twocolumn,letterpaper]{article}

\usepackage[pagebackref=true,breaklinks=true,colorlinks,bookmarks=false]{hyperref}

\usepackage{cvpr}
\usepackage{times}
\usepackage{epsfig}
\usepackage{graphicx}
\usepackage{amsmath}
\usepackage{amssymb}
\usepackage{enumitem}
\usepackage{booktabs}
\usepackage{xcolor}
\usepackage{multirow}
\usepackage{subcaption}

\cvprfinalcopy 

\def\cvprPaperID{xxxx} 

\ifcvprfinal\fi

\newcommand*{\affmark}[1][*]{\textsuperscript{#1}}

\begin{document}

\title{Toward Driving Scene Understanding: \\ A Dataset for Learning  Driver Behavior and Causal Reasoning}

\author{Vasili Ramanishka\affmark[1]\\
{\tt\small vram@bu.edu}
\and
Yi-Ting Chen\affmark[2]\\
{\tt\small ychen@honda-ri.com}
\and
Teruhisa Misu\affmark[2]\\
{\tt\small tmisu@honda-ri.com}
\and
Kate Saenko\affmark[1]\\
{\tt\small saenko@bu.edu}
\and
\affmark[1]Boston University, \affmark[2]Honda Research Institute USA
}

\maketitle
\thispagestyle{empty}

\begin{abstract}
Driving Scene understanding is a key ingredient for intelligent transportation systems. To achieve systems that can operate in a complex physical and social environment, they need to understand and learn how humans drive and interact with traffic scenes. We present the Honda Research Institute Driving Dataset (HDD), a challenging dataset to enable research on learning driver behavior in real-life environments. The dataset includes 104 hours of real human driving in the San Francisco Bay Area collected using an instrumented vehicle equipped with different sensors. We provide a detailed analysis of HDD with a comparison to other driving datasets. A novel annotation methodology is introduced to enable research on driver behavior understanding from untrimmed data sequences. 
As the first step, baseline algorithms for driver behavior detection are trained and tested to demonstrate the feasibility of the proposed task.

\end{abstract}

\section{Introduction}
\begin{figure}[ht]
\begin{center}
\includegraphics[width=1.0\linewidth]{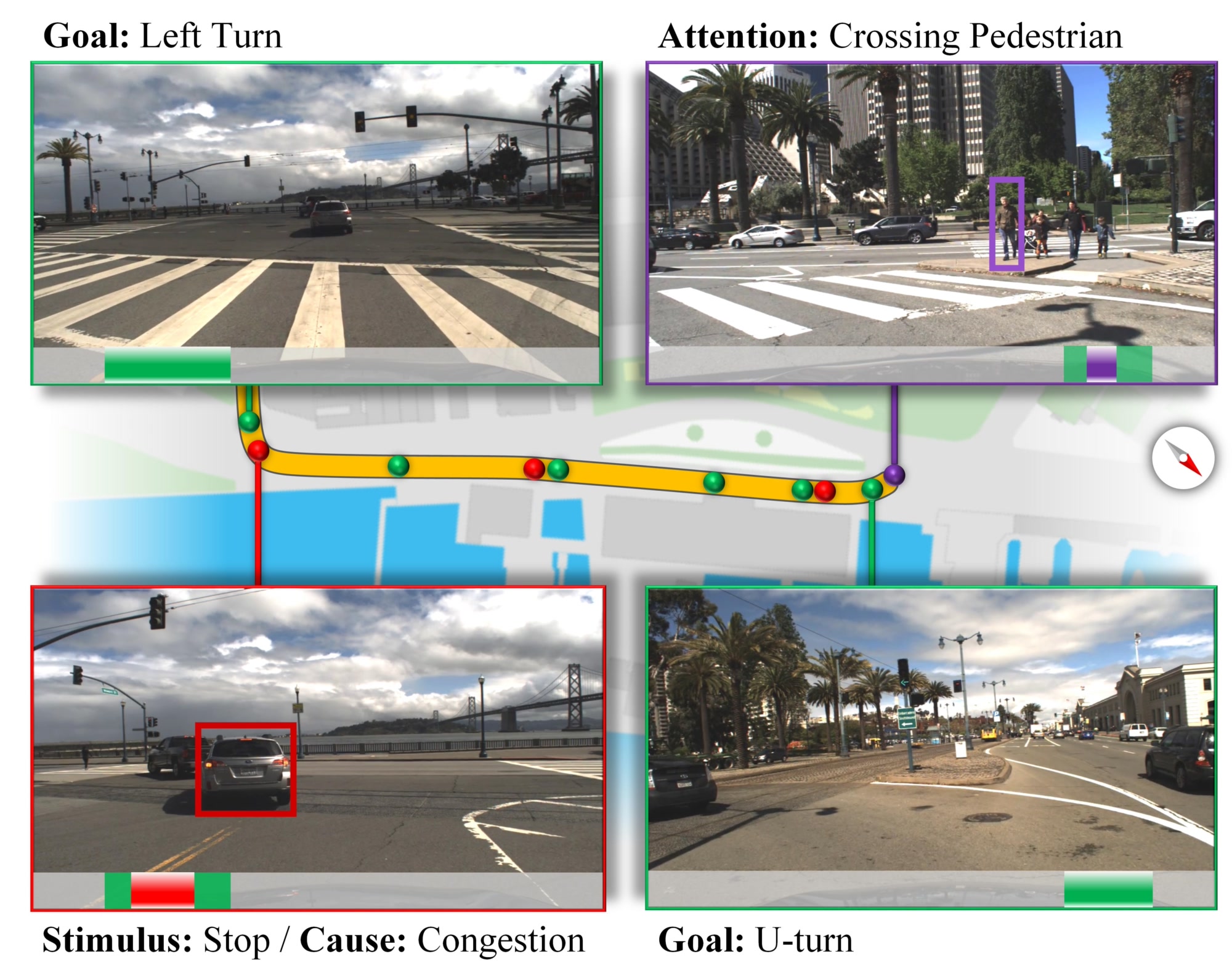}
\end{center}
\vspace{-5mm}
   \caption{\small An example illustrating different driver behaviors in traffic scenes. The yellow trajectory indicates GPS positions of our instrumented vehicle. The driver performs  actions and reasons about the scenes. To understand driver behavior, we define a 4-layer annotation scheme: \textbf{Goal-oriented action}, \textbf{Stimulus-driven action}, \textbf{Cause} and \textbf{Attention}. In \textbf{Cause} and \textbf{Attention}, we use bounding boxes to indicate when the traffic participant causes a \textit{stop} or is attended by the driver. Best viewed in color.}
\label{fig:figure1}
\vspace{-3mm}
\end{figure}

Driving involves different levels of scene understanding and decision making, ranging from detection and tracking of traffic participants, localization, scene recognition, risk assessment based on prediction and causal reasoning, to interaction.
The performance of visual scene recognition tasks has been significantly boosted by recent advances of deep learning algorithms~\cite{RenNIPS2015,HeCVPR2016,zhao2017pspnet,HeICCV2017}, and an increasing number of benchmark datasets~\cite{imagenet_cvpr09,LinECCV2014_COCO}. However, to achieve an intelligent transportation system, we need a higher level understanding.

Different vision-based datasets for autonomous driving~\cite{Geiger2012CVPR,JainICCV2015,MaddernIJRR2016,Santana2016LearningAD,CordtsCVPR2016Cityscapes,XuCVPR2017,Madhavan:EECS-2017-113} have been introduced and push forward the development of core algorithmic components. 

In core computer vision tasks, we have witnessed significant advances in object detection and semantic segmentation because of large scale annotated datasets~\cite{imagenet_cvpr09, Geiger2012CVPR,CordtsCVPR2016Cityscapes}. Additionally, the Oxford RobotCar Dataset~\cite{MaddernIJRR2016} addresses the challenges of robust localization and mapping under significantly different weather and lighting conditions. However, these datasets do not address many of the challenges in the higher level driving scene understanding. We believe detecting traffic participants and parsing scenes into the corresponding semantic categories is only the first step. Toward a complete driving scene understanding, we need to understand the interactions between human driver behaviors and the corresponding traffic scene situations~\cite{SchmidtITSC2014}. 

To achieve the goal, we design and collect HDD\footnote{The dataset is available at  \url{https://usa.honda-ri.com/HDD}} 
with the explicit goal of learning how humans perform actions and interact with traffic participants. We collected 104 hours of real human driving in the San Francisco Bay Area using an instrumented vehicle. The recording consists of 137 sessions, and each session represents a navigation task performed by a driver. Further details about the dataset will be discussed in Section~\ref{label:data_collection}. 

In each session, we decompose the corresponding navigation task into multiple predefined driver behaviors. Figure~\ref{fig:figure1} illustrates the decomposition of a navigation task. The yellow trajectory indicates GPS positions of our instrumented vehicle. A 4-layer annotation scheme is introduced to describe driver behaviors. The first layer is \textbf{Goal-oriented action}, colored green. In this example, the driver is making a \textit{left turn} as shown in the upper left image of Figure~\ref{fig:figure1}. The second layer is \textbf{Stimulus-driven action}, colored red and is shown in the lower left image. In this example, the driver makes a \textit{stop} because of \textit{stopped car}. The \textit{stop} action corresponds to \textbf{Stimulus-driven action} layer and \textit{congestion} belongs to \textbf{Cause}, which is designed to indicate the reason the vehicle makes a \textit{stop}. The red bounding box localizes the \textbf{Cause}.

While driving, human drivers are aware of surrounding traffic participants. We define the fourth layer called \textbf{Attention}, colored purple. In this example, the purple bounding box is used to indicate the traffic participant attended by the driver. Note that our annotation scheme is able to describe multiple scenarios happening simultaneously. In Figure~\ref{fig:figure1}, two different scenarios are illustrated. First, the driver intends to make a left turn but stops because of congestion. Second, the driver is making a U-turn while paying attention to a crossing pedestrian. A detailed description of our annotation methodology is presented in Section~\ref{subsec:annotation_method}.

With the multimodal data and annotations, our dataset enables the following challenging and unique research directions and applications for intelligent transportation systems. First, detecting unevenly (but naturally) distributed driver behaviors in untrimmed videos is a challenging research problem. Second, interactions between drivers and traffic participants can be explored from cause and effect labels. Third, a multi-task learning framework for learning driving control can be explored. With the predefined driver behavior labels, the annotations can be used as an auxiliary task (i.e., classification of behavior labels) to improve the prediction of future driver actions. Fourth, a multimodal fusion for driver behavior detection can be studied. 

Learning how humans drive and interact with traffic scenes is a step toward intelligent transportation systems. In this paper, we start from a driver-centric view to describe driver behaviors. However, driving involves other aspects. Particularly, it involves predicting traffic participants' intentions and reasoning overall traffic situations for motion planning and decision making, which is not discussed in this work. Toward the goal of developing intelligent driving systems, a scalable approach for constructing a dataset is the next milestone.

\section{Related work}
We review a range of datasets and highlight the uniqueness and the relationship between the current datasets and the proposed dataset.
\begin{table*}
\small
\vspace{-3mm}
\caption{Comparison of driving scene datasets}
\vspace{-6mm}

\begin{center}
\begin{tabular}{|c|c|c|c|c|}
\hline
Dataset & Purpose & Sensor types & Hours & Areas \\
\hline\hline
\begin{tabular}{@{}c@{}}Princeton \\ DeepDriving\end{tabular}~\cite{ChenICCV2015} & Vision-based control  & Driving game TORCS & ~4.5 & Driving game TORCS\\
\hline
KITTI~\cite{Geiger2012CVPR} & \begin{tabular}{@{}c@{}} Semantic understanding \& \\vision-based control~\cite{LeeCVCTC17} \end{tabular} & \begin{tabular}{@{}c@{}}Camera, LiDAR, GPS, \\ and IMU\end{tabular} & ~1.4 &  Suburban, urban and highway\\
\hline
BDD-Nexar~\cite{Madhavan:EECS-2017-113} & \begin{tabular}{@{}c@{}c@{}}Vision-based control \& \\semantic understanding \& \\ representation learning using videos\end{tabular} & Camera, GPS and IMU & 1000 &  Suburban, urban and highway\\
\hline
Udacity~\cite{UdacityDataset} & \begin{tabular}{@{}c@{}}Steering angle prediction \& \\ image-based localization\end{tabular}  & \begin{tabular}{@{}c@{}}Camera, LiDAR, GPS, \\ IMU, and CAN\end{tabular} & 8 & Urban and highway\\
\hline
comma.ai~\cite{Santana2016LearningAD} & Driving simulator  & \begin{tabular}{@{}c@{}}Camera, GPS, IMU, \\ and CAN\end{tabular} & 7.25 & Highway\\
\hline
Brain4Car~\cite{JainICCV2015} & Driver Behavior Anticipation  & \begin{tabular}{@{}c@{}}Camera, GPS,\\ and speed logger\end{tabular} & \begin{tabular}{@{}c@{}}N/A \\(1180 mi)\end{tabular} & Suburban, urban and highway\\
\hline
Ours & \begin{tabular}{@{}c@{}}Driver behavior \& \\ causal reasoning\end{tabular} & \begin{tabular}{@{}c@{}}Camera, LiDAR, GPS, \\ IMU and CAN\end{tabular}  & 104 & Suburban, urban and highway\\
\hline
\end{tabular}
\end{center}
\label{tbl:comparison_driving_dataset}
\end{table*}

{\flushleft \bf Driving Scene Datasets.}
The emergence of driving scene datasets has accelerated the progress of visual scene recognition for autonomous driving. KITTI~\cite{Geiger2012CVPR} provides a suite of sensors including cameras, LiDAR and GPS/INS. They launch different benchmarks (e.g., object detection, scene flow and 3D visual odometry) to push forward the algorithmic developments in these areas. 

Cordts et al.~\cite{CordtsCVPR2016Cityscapes} proposed a large scale road scene dataset, Cityscapes Dataset, with 5000 images with fine pixel-level semantic labeling. It enables research in category-level and instance-level semantic segmentation~\cite{zhao2017pspnet,HeICCV2017} in driving scenes that KITTI dataset does not address. For long-term localization and mapping, the Oxford Robotcar dataset~\cite{MaddernIJRR2016} presents a huge data collection collected under a variety of weather and lighting conditions over a year. 

Our dataset is complementary to~\cite{Geiger2012CVPR} and~\cite{MaddernIJRR2016} since we focus on learning driver behavior under various traffic situations. A joint effort of ours and these existing datasets can lead to intelligent transportation systems.

Recently, learning a vision-based driving model~\cite{ChenICCV2015,JainICCV2015,Santana2016LearningAD,XuCVPR2017,Jayaraman2017} for autonomous driving has attracted a lot of attention. Chen et al.~\cite{ChenICCV2015} used the driving game TORCS to obtain training data for learning a driving model by defining different affordance indicators. Jain et al.,~\cite{JainICCV2015} proposed a dataset and algorithms to anticipate driver maneuvers. Santana and Hotz~\cite{Santana2016LearningAD} presented a dataset with 7.25 hours of highway driving data to support research in this task. Earlier developments are constrained by limited amount of real-world driving data or simulated environment data~\cite{ChenICCV2015}. With these limitations in mind, the BDD-Nexar dataset~\cite{XuCVPR2017,Madhavan:EECS-2017-113}, which includes video sequences, GPS and IMU, was proposed and adopted a crowdsourcing approach to collect data from multiple vehicles across three different cities in the US. 

The proposed dataset provides additional annotatThe ions to describe common driver behaviors in driving scenes while existing datasets only consider turn, go straight, and lane change. Moreover, CAN signals are captured to provide driver behaviors under different scenarios, especially interactions with traffic participants. 

Recently, Xu et al.~\cite{XuCVPR2017} proposed an end-to-end FCN-LSTM network for this task. They considered 4 discrete actions in learning a driving model. The definition of four actions is based on CAN signals with heuristics. Instead, we provide an explicit definition of driver behaviors as shown in Figure~\ref{fig:label_distribution}. Multitask learning frameworks for learning a driving model by introducing auxiliary tasks (i.e., classification of current driver behavior and prediction of multisensor values) can be designed. A similar idea is proposed in~\cite{XuCVPR2017} in that they introduced semantic segmentation as a side task. Integrating a behavior classification task can make the models explainable to humans, and can allow the use of common sense in traffic scenes from human priors.   

A detailed comparison of different driving scene datasets is shown in Table~\ref{tbl:comparison_driving_dataset}.    

{\flushleft \bf Human Activity Understanding Datasets.}
Human activity understanding plays an important role in achieving intelligent systems. Different datasets have been proposed~\cite{caba2015activitynet,sigurdsson2016hollywood,KayKinetics2017,nakamura2017ego} to address the limitations in earlier works. Note that our data can enable research in learning driver behaviors as mentioned in the introduction. Particularly, recognizing a \textbf{Goal-oriented action} is an \textit{egocentric activity recognition} problem. The Stanford-ECM dataset~\cite{nakamura2017ego} is related to our dataset in the following two aspects. First, they define egocentric activity classes for humans as in our \textbf{Goal-oriented} classes for drivers. Second, they provide egocentric videos and signals from a wearable sensor for jointly learning activity recognition and energy expenditure while we provide multisensor recordings from an instrumented vehicle for learning driver behavior. In addition to learning egocentric activities, we also  annotate how traffic participants interact with drivers.   

Datasets for research on pedestrian behaviors are released~\cite{KooijECCV2014,RasouliICCVW2017}. Kooij et al.,~\cite{KooijECCV2014} proposed a dataset that annotates a pedestrian with the intention to cross the street under different scenarios. Rasouli et al.,~\cite{RasouliICCVW2017} provides a dataset (Joint Attention in Autonomous Driving) with annotations for studying pedestrian crosswalk behaviors. Understanding interactions between the driver and pedestrian are important for decision making. Robust pedestrian behavior modeling is also necessary~\cite{KitaniECCV2012,KellerITS2014}.

{\flushleft \bf Visual Reasoning Datasets.}
Visual question answering (VQA) is a challenging topic in artificial intelligence. A VQA agent should be able to reason and answer questions from visual input. An increasing number of datasets~\cite{AgrawalICCV2015,JohnsonCVPR2017} and algorithms~\cite{AndreasCVPR2016,HuICCV2017} have been proposed recent years. Specifically, CLEVR dataset~\cite{JohnsonCVPR2017} is presented to enable the community to build a strong intelligent agent instead of solving VQA without reasoning. In our work, we hope to enable the community to develop systems that can understand traffic scene context, perform reasoning and make decisions.

\section{Honda Research Institute Driving Dataset}\label{label:data_collection}
\subsection{Data Collection Platform }
\begin{figure}
\begin{center}
\includegraphics[width=0.8\linewidth]{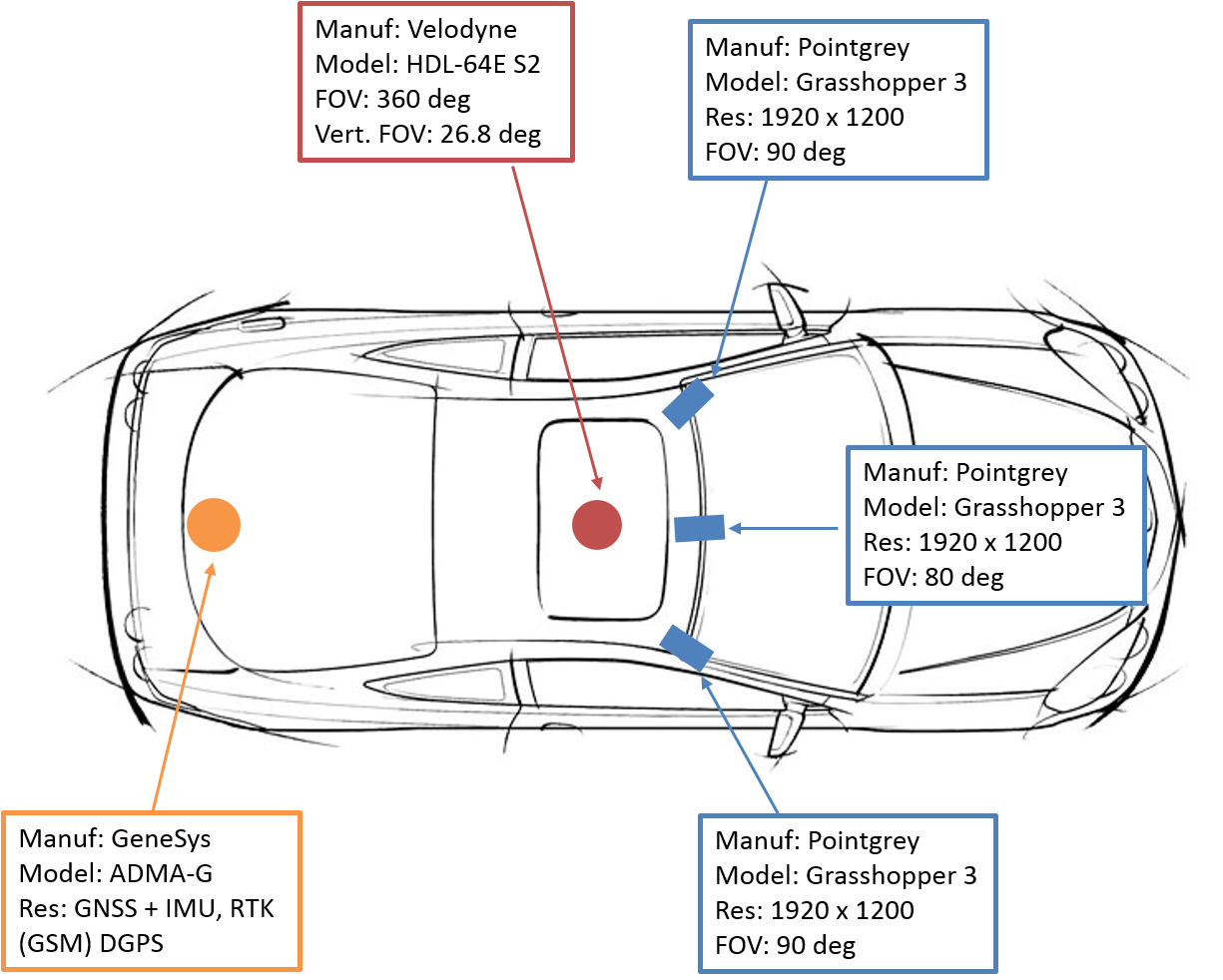}
\end{center}
\vspace{-5mm}
   \caption{Sensor layout of the instrumented vehicle.}
\vspace{-3mm}
\label{fig:sensor_layout}
\end{figure}

The data was collected using an instrumented vehicle equipped with the following sensors (their layout is shown in Figure~\ref{fig:sensor_layout}):

\begin{enumerate}[label=(\roman*)]
\small
\itemsep0em 
\item 3 x Point Grey Grasshopper 3 video camera, resolution: 1920 × 1200 pixels, frame rate: 30Hz, field of view (FOV): 80 degrees x 1 (center) and 90 degrees x 2 (left and right).
\item 1 x  Velodyne HDL-64E S2 3D LiDAR sensor, spin rate: 10 Hz, number of laser channel: 64, range: 100 m, horizontal FOV: 360 degrees, vertical FOV: 26.9 degrees.
\item 1 x GeneSys Eletronik GmbH Automotive Dynamic Motion Analyzer with DGPS outputs gyros, accelerometers and GPS signals at 120 Hz.
\item a Vehicle Controller Area Network (CAN) that provides various signals from around the vehicle. We recorded throttle angle, brake pressure, steering angle, yaw rate and speed at 100 Hz. 
\end{enumerate}

All sensors on the vehicle were logged using a PC running Ubuntu Linux 14.04 with two eight-core Intel i5-6600K 3.5 GHz Quad-Core processors, 16 GB DDR3 memory, and a RAID 0 array of four 2TB SSDs, for a total capacity of 8 TB. The sensor data are synchronized and timestamped using ROS\footnote{http://www.ros.org/} and a customized hardware and software designed for multimodal data analysis. 

\begin{figure}
\begin{center}
\frame{\includegraphics[width=0.9\linewidth, trim={0.5cm 13cm 0.5cm 0.3cm},clip]{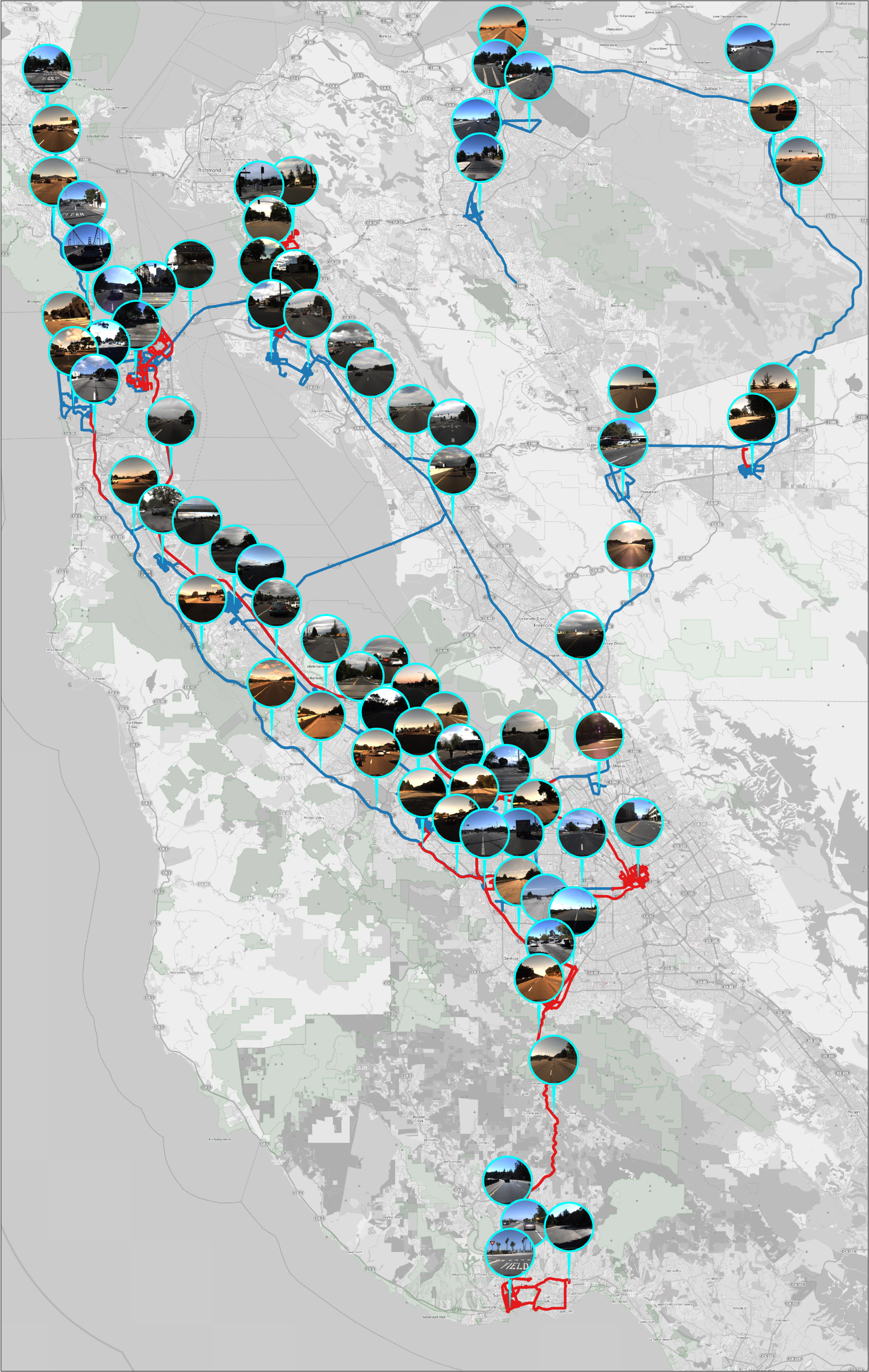}}
\end{center}
\vspace{-5mm}
\caption{\small The figure shows GPS traces of the HDD dataset. We split the dataset into training and testing according to the vehicle's geolocation. The blue and red color traces denote the training and testing sets, respectively.}
\vspace{-3mm}
\label{fig:split_map}
\end{figure}

For our dataset, we are interested in having a diverse set of traffic scenes with driver behaviors. The  current data collection spans from February 2017 to October 2017. We drove within the San Francisco Bay Area including on urban, suburban and highway roads, as shown in Figure~\ref{fig:split_map}. The total size of the post-processed dataset is around 150 GB and 104 video hours. The video is converted to a resolution of 1280 $\times$ 720 at 30 fps.  

\label{subsec:annotation_method}
\begin{figure}
\begin{center}
\includegraphics[width=0.92\linewidth]{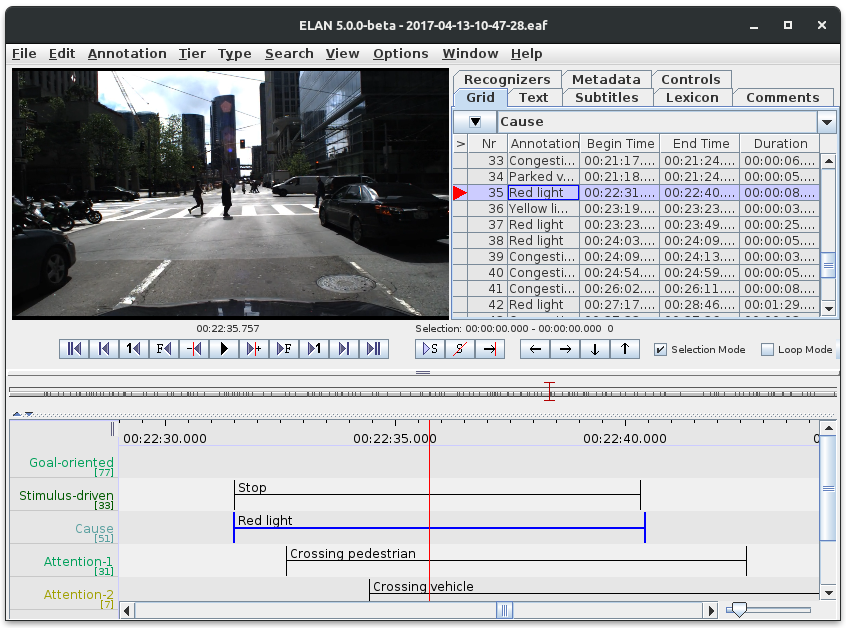}
\end{center}
\vspace{-5mm}
   \caption{\small Annotation interface. We use the open source software toolkit ELAN to annotate different driver behaviors and causal relationships.}
\vspace{-5mm}
\label{fig:annotation_interface}
\end{figure}

\begin{figure*}
\includegraphics[width=0.99\textwidth]{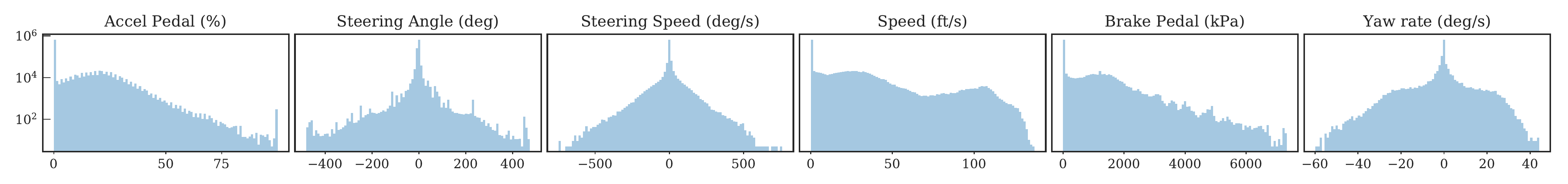}
\vspace{-10pt}
\caption{Histograms of sensor measurements for the dataset.}
\label{fig:sensor_distribution}
\vspace{-10pt}
\end{figure*}

\subsection{Annotation Methodology}
It is challenging to define driver behavior classes since it involves cognitive processes and vehicle-driver interaction. It is especially challenging to identify an exact segment of driver behavior from data we collected, in particular from video sequences. In our annotation processes, we make the best effort in annotating different driver behaviors with a mixture of objective criteria and subjective judgment. 

Our annotation methodology is motivated by human factor and cognitive science. Michon~\cite{MichonHBTS1985} proposed three classes of driving processes:  \textbf{operational processes} that correspond to the manipulation of the vehicle, \textbf{tactical processes} that are the interactions between the vehicle, traffic participants and environment, and \textbf{strategic processes} for higher level reasoning, planning and decision making. 

With these definitions in mind, we propose a 4-layer representation to describe driver behavior, i.e., \textbf{Goal-oriented action}, \textbf{Stimulus-driven action}, \textbf{Cause} and \textbf{Attention}, which encapsulate driver behavior and causal reasoning. A complete list of labels in the 4-layer representation for describing driver behavior can be found in Figure~\ref{fig:label_distribution}. 

\textbf{Goal-oriented action} involves the driver's manipulation of the vehicle in a navigation task such as \textit{right turn}, \textit{left turn}, \textit{branch} and \textit{merge}. While operating the vehicle, the driver can make a \textit{stop} or \textit{deviate} due to traffic participants or obstacles. \textit{Stop} and \textit{deviate} are categorized as \textbf{Stimulus-driven action}. When the driver performs a \textit{stop} or a \textit{deviate} action, there is a reason for it. We define the third layer \textbf{Cause} to explain the reason for these actions. For example, a \textit{stopped car} in front of us is an immediate cause for a stop as in Figure~\ref{fig:figure1}. Finally, the fourth layer \textbf{Attention} is introduced to localize the traffic participants that are attended by drivers. For example, a \textit{pedestrian near ego lane} may be attended by drivers since the pedestrian might perform certain actions that would affect driver behavior.   

Based on the aforementioned definition, we work with experienced external annotators on this task. The annotators use an open source software package ELAN\footnote{https://tla.mpi.nl/tools/tla-tools/elan/} to label videos as shown in Figure~\ref{fig:annotation_interface}. To ensure the consistency in annotations, we conduct the following quality control strategy. Given a driving session, it is first annotated by 2 independent human annotators. Then, a third annotator merges the three annotations with his/her own judgment into a single annotation. Finally, we have an internal expert annotator to review and obtain the final version.

To analyze the annotation consistency, we compare the annotation quality of the third external annotator to the internal expert annotator on 10 sessions. Based on the same annotation procedures, we found a 98\% (driver behavior label) agreement between the third external annotator and the internal expert annotator. However, the start time and end time of a driver behavior is not trivial to assess since it involves a subjective judgment. A systematic evaluation of action localization consistency is needed and requires a further investigation.   

\subsection{Dataset Statistics}
\begin{figure*}

\begin{subfigure}[b]{1\textwidth}
\begin{center}
\includegraphics[width=\linewidth, trim={0.1cm 0cm 0.1cm 0},clip]{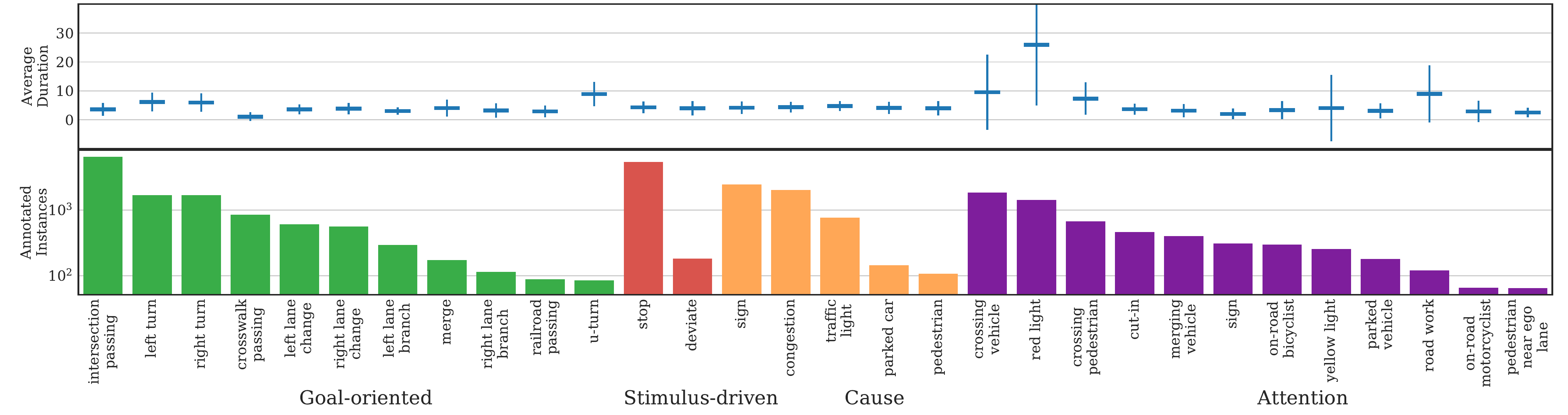}
\caption{Driver behavior statistics (annotated instances and instance duration) in different layers.}
\label{fig:label_distribution}
\end{center}
\end{subfigure}

\begin{center}
\vspace{-6pt}
\begin{subfigure}[t]{0.52\textwidth}
   \includegraphics[width=1\linewidth, trim={0.1cm 0cm 0.3cm 0},clip]         {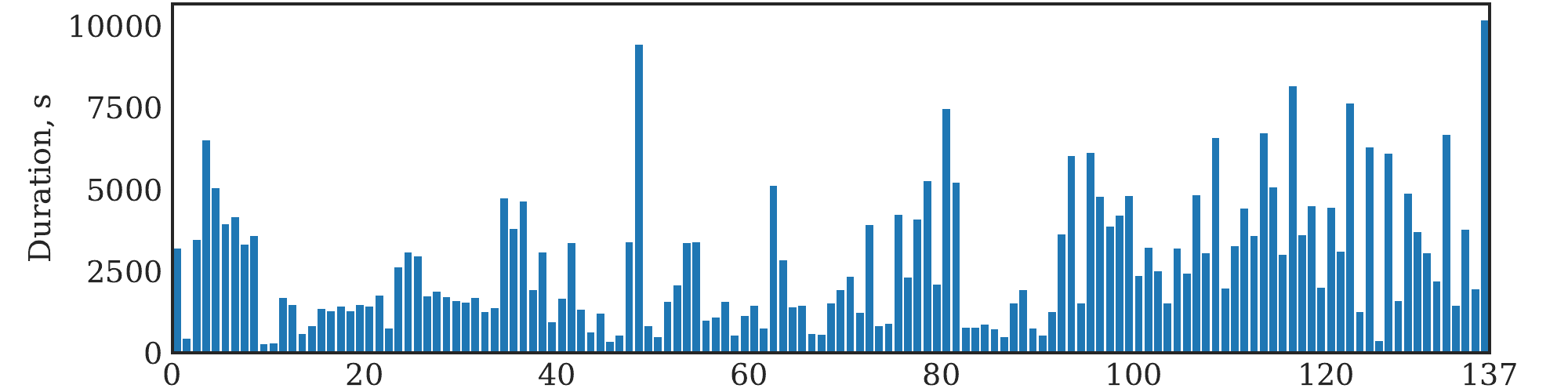}
   \caption{A total of 137 annotated sessions with a 45-minute average duration.}
   \label{fig:average_duration}
\end{subfigure}\hfill
\begin{subfigure}[t]{0.45\textwidth}
   \includegraphics[width=1\linewidth, trim={0.5cm 26.9cm 13cm 0.5cm}, clip]{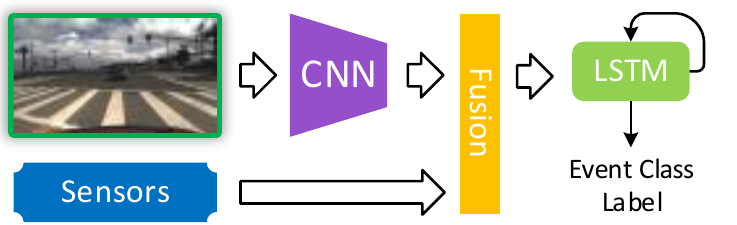}
   \caption{A baseline architecture of driver behavior detection.}
   \label{fig:model_figure1} 
\end{subfigure}
\begin{subfigure}[b]{1\textwidth}
\raggedleft
\vspace{-2pt}
   \includegraphics[width=\linewidth, trim={0.5cm 0.5cm 0.5cm 1.5cm}, clip]{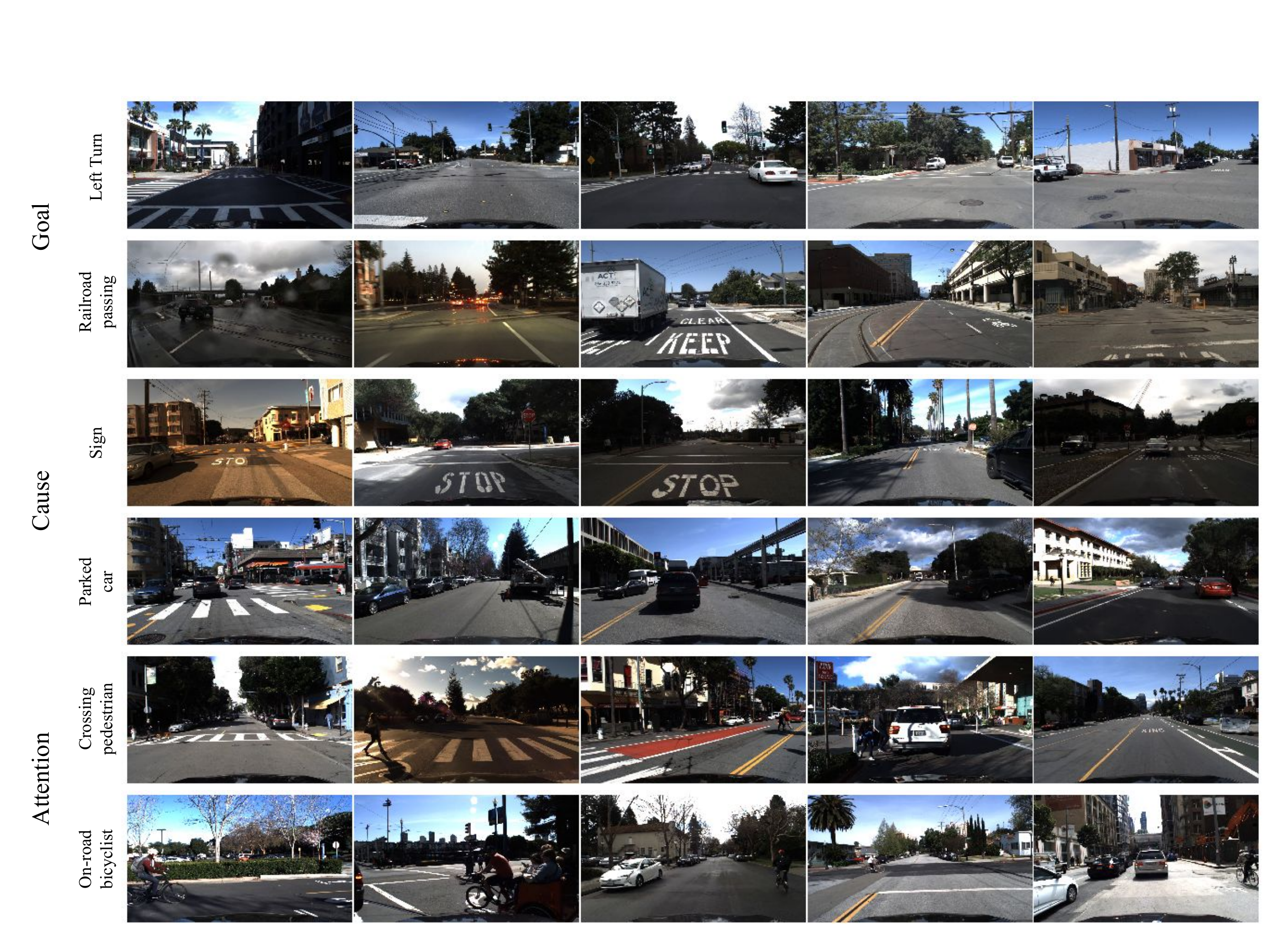}
   \caption{Examples of video frames from our dataset which belong to driver behaviors designated on the left.}
   \label{fig:dataset_examples} 
\end{subfigure}
\end{center}

\vspace{-10pt}
\caption{}
\label{fig:sensorsValues}

\end{figure*}

In the current release, we have a total of 104 video hours, which are annotated with the proposed 4-layer structure. Within 104 video hours, we have 137 sessions corresponding to different navigation tasks. The average duration of each session is 45 minutes. The statistics of session duration can be found in Figure~\ref{fig:average_duration}. Figure~\ref{fig:label_distribution} shows the number of instances of each behavior. A highly imbalanced label distribution can be observed from the figure.

\begin{table*}
\footnotesize
\centering
\caption{\small Detection results on test set for \textbf{Goal-oriented action} using per-frame average precision, \textit{Random} illustrates the portion of frames which belongs to the given action. \textit{Sensors} model uses LSTM to encode the stream of 6-dimensional vectors from CAN bus. Features from the last convolutional layer of InceptionResnet-V2 on RGB frames are provided as an input to LSTM in \textit{CNN} models.}
\vspace{-6pt}
\label{table:results}
\begin{tabular}{l||c|c|@{\,}c@{\,}|c|c|c|c|c|@{\,}c@{\,}|c|c||c}
Models &
\begin{tabular}{@{}c@{}}right \\ turn\end{tabular} & 
\begin{tabular}{@{}c@{}}left \\ turn\end{tabular} &
\begin{tabular}{@{}c@{}}intersection \\ passing\end{tabular} & 
\begin{tabular}{@{}c@{}}railroad \\ passing\end{tabular} &
\begin{tabular}{@{}c@{}}left \\ lane \\ branch\end{tabular} &
\begin{tabular}{@{}c@{}}right \\ lane \\ change\end{tabular} &
\begin{tabular}{@{}c@{}}left \\ lane \\ change\end{tabular} &
\begin{tabular}{@{}c@{}}right \\ lane \\ branch\end{tabular} &
\begin{tabular}{@{}c@{}}crosswalk \\ passing \end{tabular} &
merge &
u-turn &
mAP   \\ \midrule
Random         & 2.24  & 2.61  & 6.58  & 0.07 & 0.15 & 0.46  & 0.42  & 0.07 & 0.21 & 0.13 & 0.18  & 1.19  \\
CNN pool       & 30.11 & 31.88 & 56.49 & 3.96 & 2.02 & 1.35  & 1.43  & 0.15 & 8.71 & 7.13 & 4.89  & 13.46 \\
Sensors        & 74.27 & 66.25 & 36.41 & 0.07 & 8.03 & 13.39 & 26.17 & 0.20 & 0.30 & 3.59 & 33.57 & 23.84 \\
CNN conv       & 54.43 & 57.79 & 65.74 & 2.56 & 25.76& 26.11 & 27.84 & 1.77& 16.08 & 4.86 & 13.65 & 26.96 \\
CNN+Sensors    & 77.47 & 76.16 & 76.79 & 3.36 & 25.47 & 23.08 & 41.97 & 1.06 & 11.87 & 4.94 & 17.61 & 32.71 \\
\bottomrule
\end{tabular}
\vspace{-8pt}
\end{table*}

\begin{table}
\footnotesize
\centering
\caption{\small Detection results on test split for \textbf{Cause layer}. Behaviors from this layer are immediate causes for either \textit{stop} or \textit{deviate} actions.} 
\label{table:results_cause}
\begin{tabular}{l||c|@{\,}c@{\,}|@{\,}c@{\,}|@{\,}c@{\,}|@{\,}c@{\,}||c}
Models &
\begin{tabular}{@{}c@{}}sign\end{tabular} & 
\begin{tabular}{@{}c@{}}congestion\end{tabular} &
\begin{tabular}{@{}c@{}}traffic\\light\end{tabular} & 
\begin{tabular}{@{}c@{}}pedestrian\end{tabular} &
\begin{tabular}{@{}c@{}}parked\\car\end{tabular} &
mAP   \\ \midrule
Random         & 2.49 & 2.73 & 1.22 & 0.20 & 0.15 & 1.36 \\
CNN+Sensors    & 46.83 & 39.72 & 45.31 & 2.15 &	7.24  & 28.25  \\
\bottomrule
\end{tabular}
\label{tbl:cause}
\vspace{-10pt}
\end{table}

\section{Multimodal Fusion for Driver Behavior Detection}
Our goal is to detect driver behaviors which occur during driving sessions by predicting a probability distribution over the list of our predefined behavior classes at every given point of time. As the first step, we focus on the detection of \textbf{Goal-oriented} and \textbf{Cause} layers in our experiments. To detect driver behaviors, we design an algorithm to learn a representation of driving state which encodes the necessary history of past measurements and effectively translates them into probability distributions. Long-Short Term Memory (LSTM) networks were shown to be successful in many temporal modeling tasks, including activity detection. We thus employ an LSTM as the backbone architecture for our model.  

In addition to the input video stream, our model has access to an auxiliary signal which provides complimentary information about the vehicle dynamics. This auxiliary signal includes measurements from the following CAN bus sensors: car speed, accelerator and braking pedal positions, yaw rate, steering wheel angle, and the rotation speed of the steering wheel, illustrated in Figure~\ref{fig:sensor_distribution}. This makes the task and approach different from the standard activity detection setup where models usually have access only to a single modality. The proposed baseline architecture of driver behavior detection is shown in Figure~\ref{fig:model_figure1}.

Our application domain dictates the necessity to design a model which can be used in real-time in a streaming manner. Thus, we constrain our model to the case where predictions are made solely based on the current input and previous observations.

\section{Experiments}
To assess the quality of the visual representations learned by our model on our challenging dataset, we perform rigorous experiments as well as comparisons to several baselines. First of all, we split the dataset based on the geolocation data, thus, minimizing spatial overlap of train and test routes. This way we avoid testing on the very same locations as those used to train the model. Fig~\ref{fig:split_map} shows the geolocation measurements in the training (in blue) and test (in red) splits.

For baseline models we sample input frames from video streams and values from CAN bus sensors at 3 Hz. We found this to be a reasonable trade-off between modeling complexity and precision. We employed the \textit{Conv2d\_7b\_1x1} layer of InceptionResnet-V2~\cite{SzegedyIV16} pretrained on ImageNet~\cite{Deng09imagenet} to get feature representations for each frame. 

Raw sensor values are passed through a fully-connected layer before concatenation with visual features. In turn, visual features are represented by spatial grid of CNN activations. Additional 1x1 convolution is applied to reduce dimensionality of \textit{Conv2d\_7b\_1x1} from $8 \times 8 \times 1536$ to $8 \times 8 \times 20$ before flattening it and concatenating with sensor features. The necessity to preserve the spatial resolution is illustrated by Table~\ref{table:results} where `CNN conv' demonstrates the substantial advantage in detection of turns. LSTM hidden state size is set to $2000$ in all experiments. During training, we formed batches of sequence segments by sequentially iterating over driving sessions. The last LSTM hidden state from the previous batch is used to initialize the LSTM hidden state on the next step. Training is performed using truncated backpropagation through time.  

For Goal and Cause layers, we trained separate LSTMs using batches of size 40 with each sequence length set to 90 samples. We confirmed that the larger batch size improves convergence. We set dropout \textit{keep} probability on the input and output of the LSTM to 0.9. Taking into account the data imbalance between foreground and background frames, and also the imbalance of behavior classes themselves, we used the recently proposed technique for modifying cross-entropy loss to deal with class imbalance~\cite{Lin2017b}. This modification  was originally applied to the task of object detection where negative region proposals also dominate.  

Our evaluation strategy is inspired from the activity detection literature~\cite{CDC17} where each frame is evaluated for the correct activity label. Specifically, Shou \textit{et al.}~\cite{CDC17} treated the per-frame labeling task as a retrieval problem and computed Average Precision (AP) for each activity class by first ranking all frames according to their confidence scores. Following this procedure, we compute the AP for individual driver behavior classes as well as the mean AP (mAP) over all behavior classes. The test split we obtained via the geospatial selection procedure described above includes 37 driving sessions, which contain a total of approximately 274,000 frames sampled at 3 FPS for the mean average precision evaluation.

\begin{figure*}
\centering
\begin{subfigure}[t]{.43\textwidth} 
\includegraphics[width=\linewidth, trim={0.5cm 1.5cm 0.2cm 0.5cm},clip]{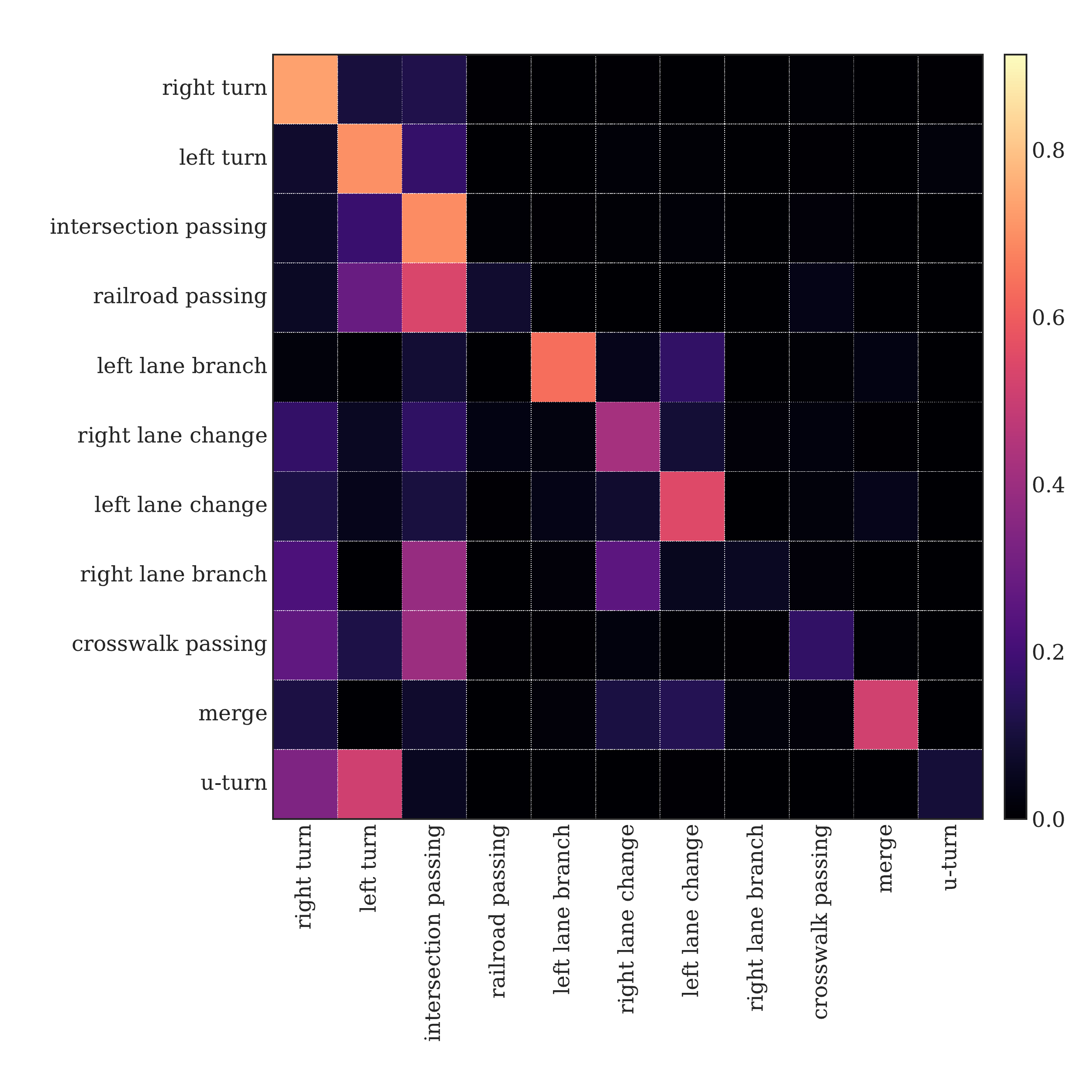}
\label{fig:confusion_matrix_1}
\end{subfigure}
~\hfill~
\begin{subfigure}[t]{.43\textwidth} 
\includegraphics[width=\linewidth, trim={0.5cm 1.5cm 0.2cm 0.5cm},clip]{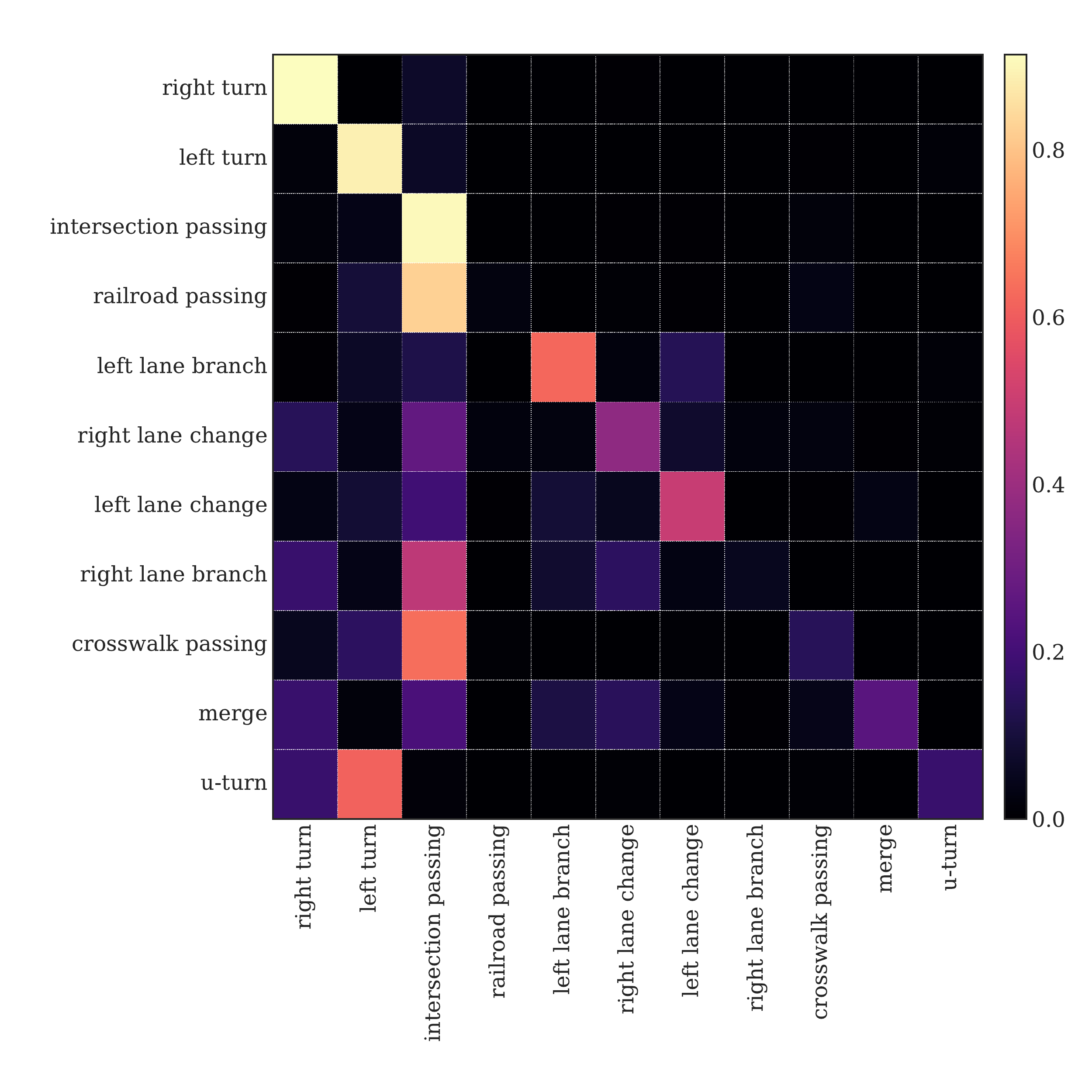}
\label{fig:confusion_matrix_2}
\end{subfigure}
\vspace{-13pt}
\caption{\small Confusion matrices for \textit{Goal-oriented} driver behavior classes using `CNN conv' model (left) and `CNN+Sensors' (right). For better view we omit background class from visualization and normalize rows accordingly. Note the clear disambiguation between the detection of turns and `intersection passing' after adding the input data from sensors. Intersections usually have crosswalks, signs and road markings which influence detection performance for other classes.}
\vspace{-4pt}
\end{figure*}

\section{Results and Discussion}

\noindent\textbf{Goal-oriented layer} Table~\ref{table:results} provides the APs for 11 Goal-oriented Actions (starting from `right turn' to `u-turn') for our model and its ablated versions. The last column provides the mAP value for all the methods. First, we provide a description of the baselines used in this experiment. The first baseline (`Random') simply assigns random behavior labels to each frame and serves as the lower bound on model performance. It also illustrates the portion of frames in the test data for which every class label is assigned. The next one (`CNN pool') encodes each frame by extracting convolutional features using an InceptionResnet-V2 network and pooling them spatially to a fixed-length vector. These pooled representations of frames are sequentially fed to the LSTM to predict the behavior label. The third baseline ('Sensors') uses only the sensor data as input to the LSTM. The next method (`CNN conv') is a variant of the second method: instead of spatially pooling CNN feature encodings, we used a small convnet to reduce the dimensionality of the CNN encodings of the frames before passing them through the LSTM. Finally, the `CNN+Sensors' method adds sensor data to the `CNN conv' method.

We can see that the performance of `CNN pool' is quite low. This can be attributed to the fact that information is lost by the spatial pooling operation. `CNN conv' replaces pooling by a learnable conv layer and significantly increases mAP. Sensor measurements (brake, steering wheel, etc.) alone result in slightly better AP for simple actions like left/right turns where the information about steering wheel position can be sufficient in most of the cases. When it comes to actions like lane changes, visual information used in `CNN conv' allows for proper scene interpretation, thus, improving over `Sensor' model. It is clear, that only sensor information is not sufficient for driver behavior detection, especially in an imbalanced scenario. The visual data and data from sensors are complementary to each other in this respect and thus their fusion gives the best results, as shown in the last row of the table.

It is interesting to note that the `blind model' (without camera input) is able to successfully guess `intersection passing' because most of them are happening in a very specific pattern: `deceleration/wait/acceleration'. A `railroad passing' is surprisingly hard for the CNN model because this behavior type includes not only railroad crossing in the designated locations which have discriminative visual features but also tram rails crossing. The confusion of behavior classes with a `background' class remains the most frequent source of errors for all layers.

\noindent\textbf{Cause Layer}
Table~\ref{tbl:cause} represents the detection results of causes for stimulus-driven actions. We observe a better detection performance for \textit{sign}, \textit{congestion} and \textit{traffic light}. The corresponding motion pattern should be similar, i.e., deceleration. On the other hand, the vehicle dynamics for \textit{pedestrian} and \textit{parked car} are very different from the rest. For \textit{pedestrian}, the vehicle usually makes a stop action for pedestrians while making turns. For \textit{parked car}, the vehicle deviates from the original trajectory to avoid a collision. We hypothesize that the weak performance of the proposed model is due to the following two reasons. First, the two classes are underrepresented in the dataset as shown in Figure~\ref{fig:label_distribution}. A better approach to deal with an imbalanced distribution is necessary. The same is true for rare \textbf{Goal-oriented} actions. Second, the motion modeling for a short duration of cause (e.g., pedestrian) may not be captured in the baseline model (similar to railroad passing). The motion pattern of a \textit{deviate} action may not be modeled effectively to detect \textit{parked car}. This would benefit from better motion modeling using optical flow features. We leave this for future work.  

In the current version of the dataset, we have only four causes for a stop action, namely, \textit{sign}, \textit{congestion}, \textit{traffic light}, \textit{pedestrian}, and one cause \textit{parked car} for a deviate action. Because detection of these immediate causes directly implies detection and discrimination of their respective actions we do not provide separate results for \textbf{Stimulus-driven} layer.

\section{Conclusion}
In this paper, we introduce the Honda Research Institute Driving Dataset, which aims to stimulate the community to propose novel algorithms to capture the driver behavior. To enable this, we propose a novel annotation methodology that decomposes driver behaviors into a 4-layer representation, i.e., \textbf{Goal-oriented}, \textbf{Stimulus-driven}, \textbf{Cause} and \textbf{Attention}.
A variety of baselines for detecting driver behaviors in untrimmed videos were proposed and tested on this dataset.
Our preliminary results show that this task is challenging for standard activity recognition methods based on RGB frames. Although adding sensor data improves accuracy, we need better representations, temporal modeling and training strategy to achieve reasonable performance in driver behavior detection before exploring the actual relationship between behaviors in different layers, i.e., the relationship between a driver and traffic situations.

\section{Acknowledgements}
This work is supported in part by the DARPA XAI program and Honda Research Institute USA.

{\small
\bibliographystyle{ieee}
\bibliography{egbib}
}
\end{document}